\documentclass[pdflatex,sn-mathphys-num]{sn-jnl}% Math and Physical Sciences Author Year Reference Style
\renewcommand{\cite}{\citep}
\usepackage{graphicx}%
\usepackage{multirow}%
\usepackage{amsmath,amssymb,amsfonts}%
\usepackage{amsthm}%
\usepackage{mathrsfs}%
\usepackage[title]{appendix}%
\usepackage[dvipsnames]{xcolor}
\usepackage{textcomp}%
\usepackage{manyfoot}%
\usepackage{booktabs}%
\usepackage{algorithm}%
\usepackage{algorithmicx}%
\usepackage{algpseudocode}%
\usepackage{listings}%
\usepackage{booktabs}
\usepackage{comment}
\usepackage{array}
\usepackage{ragged2e}
\usepackage{makecell}
\usepackage{multirow}
\usepackage{epigraph}

\usepackage{tabularx}
\usepackage{ragged2e}
\newcolumntype{L}{>{\raggedright\arraybackslash}X}

\usepackage{lmodern}

\usepackage[colorinlistoftodos,textsize=scriptsize,textwidth=2cm]{todonotes}

\usepackage{xspace}
\usepackage{quoting}
\usepackage{comment}

\theoremstyle{thmstyleone}%
\theoremstyle{thmstyletwo}%

\theoremstyle{thmstylethree}%

\newcommand{\crterm}{cultural intelligence\xspace}

\usepackage{booktabs}

\usepackage{tabularray}

\usepackage{array} % for \newcolumntype macro
\newcolumntype{N}{% increment and display the value of 'rownum' counter
   >{\refstepcounter{rownum}\therownum}l}
\usepackage{tabularx}
\usepackage{ragged2e}
\usepackage{geometry}
\usepackage{amsmath}

\newcolumntype{L}{>{\raggedright\arraybackslash}X}

\begin{document}

% \title[Measuring Cultural Intelligence]{Measuring Cultural Intelligence of AI Systems:\\ A Unified Framework}
\title[Quantifying Cultural Intelligence]{A Unified Framework to Quantify Cultural Intelligence of AI}

% \title[Cultural Relevance Benchmark]{A Unified Framework for Estimating Model Cultural Intelligence}


% \author*{\fnm{Atlas Cultural Benchmark} \sur{Team}}\email{do-not-email@google.com}
% \begin{comment}

 \author{\fnm{Sunipa} \sur{Dev}}\email{sunipadev@gmail.com}
\equalcont{These authors contributed equally to this work.}

 \author{\fnm{Vinodkumar} \sur{Prabhakaran}}\email{vinodkpg@google.com}
 \equalcont{These authors contributed equally to this work.}

\author{\fnm{Rutledge} \sur{Chin Feman}}

 \author{\fnm{Aida} \sur{Davani}}
 
  \author{\fnm{Remi } \sur{Denton}} 
  
  \author{\fnm{Charu} \sur{Kalia}}
  \author{\fnm{Piyawat} \sur{Lertvittayakumjorn}}
  \author{\fnm{Madhurima} \sur{Maji}}
  \author{\fnm{Rida} \sur{Qadri}}
  \author{\fnm{Negar} \sur{Rostamzadeh}}
  \author{\fnm{Renee} \sur{Shelby}}
  \author{\fnm{Romina} \sur{Stella}}
  \author{\fnm{Hayk} \sur{Stepanyan}}
  \author{\fnm{Erin} \sur{van Liemt}}
  \author{\fnm{Aishwarya} \sur{Verma}}
  \author{\fnm{Oscar} \sur{Wahltinez}}
  \author{\fnm{Edem} \sur{Wornyo}}
  \author{\fnm{Andrew} \sur{Zaldivar}}
  \author{\fnm{Saška} \sur{Mojsilović}}

\affil{Google Research}
% \end{comment}


%\affil{\orgdiv{Google Research}}
% , \orgname{Organization}, \orgaddress{\street{Street}, \city{City}, \postcode{100190}, \state{State}, \country{Country}}}

% \affil[2]{\orgdiv{Department}, \orgname{Organization}, \orgaddress{\street{Street}, \city{City}, \postcode{10587}, \state{State}, \country{Country}}}

% \affil[3]{\orgdiv{Department}, \orgname{Organization}, \orgaddress{\street{Street}, \city{City}, \postcode{610101}, \state{State}, \country{Country}}}


%\linenumbers

\abstract{
As generative AI technologies are increasingly being launched across the globe, assessing their competence to operate in different cultural contexts is urgently 
becoming a priority. 
While recent benchmarking efforts have made revealed significant gaps in cultural competence of AI models, they remain fragmented, often focusing on specific cultural aspects or evaluation tasks. 
This fragmentation hinders the ability of model developers, users, and policymakers to comprehensively assess the readiness of these models to be deployed or used in specific regional or cultural environments.
To bridge this gap, we propose a unified, structured, and fine-grained framework for assessing cultural intelligence at scale.
Drawing on psychometric measurement theory, our approach decouples the conceptual definition of cultural intelligence from its operationalization.
We start by developing a working definition of culture that includes identifying core domains, topics, and facets of culture. 
We then conceptualize cultural intelligence as a suite of latent capabilities spanning diverse domains, and demonstrate operationalizing these capabilities through a series of measurable indicators designed for reliability and extensibility. Finally, we identify the practical considerations and research pathways for data and metrics, providing a scalable roadmap for the comprehensive and responsible assessment of AI cultural competence.

}



\maketitle

\clearpage
\section{Introduction}
\label{sec_intro}
Artificial intelligence (AI) has rapidly transitioned from a niche technology to a ubiquitous component of our daily lives. Millions now interact with generative AI systems for applications ranging from seeking information \cite{skjuve2024people}, to complex problem-solving \cite{didolkar2024metacognitive}, to companionship \cite{liu2024chatbot}, while specialized AI models are reshaping core knowledge ecosystems in domains such as health \cite{zhang2023generative}, education \cite{zhang2024systematic}, and art \cite{zhou2024generative}. These advancements are also accelerating from generative tools toward more autonomous agentic tools \cite{acharya2025agentic,schneider2025generative}, with capabilities claimed to be approaching general intelligence \cite{bubeck2023sparks}. 
While the benchmark-driven development \cite{
bommasani2023holistic,srivastava2023beyond} has been instrumental in advancing AI performance, especially in logic, math, language, and reasoning, existing evaluations often overlook whether a model can truly navigate the contextual sensitivities inherent to diverse global cultures. This gap becomes critical as these models are increasingly deployed in social, and often sensitive, roles such as companionship and education.
Models can exhibit a superficial, yet convincing, illusion of cultural intelligence, for instance, by conversing fluently in a local language, or responding to certain culture-laden requests, 
but without reliable culturally grounded reasoning, thus
creating a false sense of competency and safety for users across cultural contexts.

By ``cultural intelligence'', we refer to 
a model's capacity to strategically adapt across diverse contexts, generating responses that are not only factually grounded in accurate cultural knowledge but also respectful of relevant norms and values
(we define this term formally in Section~\ref{sec_cultural_intel_defn}).
Failing to assess and improve cultural intelligence in AI may give rise to a multitude of risks. 
The first is a \textit{functional risk}: models cannot be truly helpful unless they are competent to serve user needs around local cultural knowledge in local languages they may prefer, while also adhering to implicit local norms (e.g., \cite{joshi-etal-2020-state,zhu2024language}). This may lead directly to both a \textit{socio-economic risk}---as performance disparities across cultural contexts will limit AI proliferation, adoption, and economic benefit in new markets \cite{bughin2018notes}, as well as an \textit{equity risk}---erecting barriers for diverse communities from accessing the benefits of technological advancements \cite{blasi2022systematic}. There is also the critical \textit{safety risk}: 
culturally incompetent models may cause unique safety failures where standard guardrails break down, overlooking offensive violations of social norms that are only detectable through a nuanced cultural lens \cite{wang2024all}. 
Finally, and perhaps most profoundly, there is the systemic \textit{ethical risk} of reproducing a dominant, hegemonic culture, that leads to the exclusion and marginalization of other worldviews, 
leading to value imposition and cultural erasure
\cite{prabhakaran2022cultural,masoud2025cultural}.

Some of these risks and failures have already been empirically demonstrated, ranging from the inability to accurately answer culturally specific general knowledge questions \cite{myung2024blendbenchmarkllmseveryday,singh2025global} or depict salient artifacts \cite{kannen2024beyond}, to critical failures in adhering to safety guardrails \cite{shen2024language,dengmultilingual}, while reinforcing dominant sub-cultures \cite{qadri2023ai} and propagating harmful cultural stereotypes \cite{mostafazadeh-davani-etal-2025-comprehensive}.
Recent benchmarks attempt to address
these issues, for instance, measuring linguistic style \cite{havaldar2025culturally}, ``everyday knowledge'' (e.g., BLEnD \cite{myung2024blendbenchmarkllmseveryday}, Global MMLU \cite{singh2025global}), and situational etiquette (e.g., NORMAD \cite{rao2025normad}). 
Empirical efforts have also probed values encoded in LLMs \cite{arora-etal-2023-probing} and vision-language models \cite{yadav2025beyond}. 
While these works are vital, they often reduce culture to isolated data points \cite{oh2025culture}, and tend to address the problem in a piecemeal fashion, examining individual components of the puzzle rather than treating cultural intelligence as a cohesive whole.
While recent surveys (e.g., \cite{liu2025culturally,wu2025incorporating}) offer useful categorizations of these efforts, they often fail to provide a unifying and systematic evaluation framework for conceptualizing and synthesizing the diverse capabilities that contribute to cultural intelligence in AI, nor do they engage with the inherent challenges in that endeavor. 
Such a framework is imperative not only for tracking progress on integrating these capabilities into the pursuit of AGI, but also for broader stakeholders such as policy makers and governance bodies to comprehensively assess AI capabilities and safety across within local cultural contexts.

Addressing this need raises a core question: how should cultural capabilities of AI systems be conceptualized, measured, and improved?
There is a multitude of reasons why this remains a grand, open, research challenge in the AI community.
First, there currently is a lack of an operational understanding or definition of culture in AI literature;
``culture'' is a complex and contested term to define and quantify \cite{williams1976}. 
Furthermore, culture is inherently pluralistic: societies differ profoundly not only in their artifacts, practices, and norms, but also in their interpretations, meanings, and associations. They also differ in salience, as the relative importance of specific aspects like religion or rituals varies dramatically across cultures.
Secondly, even with a working definition of culture, we currently lack a systematic approach towards quantifying and evaluating cultural intelligence. A comprehensive evaluation of cultural intelligence should consider its various facets, across diverse use cases that end-users use AI models for.
Finally, even with a comprehensive set of evaluation criteria, we lack broad-coverage and high-quality data that can ground and enable large-scale evaluation and improvement of cultural intelligence, across locales, languages, and modalities.

In this paper, we introduce a broad-purpose, structured, and fine-grained framework for assessing cultural intelligence of AI systems.
Adopting the four-level measurement approach of \citet{adcock2001measurement} rooted in psychometric measurement validity theory, we explicitly decouple the background concept (i.e., cultural intelligence) from its operationalization.
We structure this process in three steps. \textit{Conceptualization}: to establish a clear basis for measurement, we first develop a systematized concept of cultural intelligence, defined in terms of a set of core capabilities that span different cultural domains. \textit{Operationalization}: 
we then operationalize each of these capabilities as an extensible set of indicators designed to reliably measure cultural intelligence, collectively. 
\textit{Measurement}: finally, 
we identify the considerations, challenges, and research pathways to meaningfully measure these indicators, specifically focusing on data collection, probing strategies, and evaluation metrics.
This structured and fine-grained framework transforms cultural intelligence into a measurable construct that is extensible and controllable in its application across diverse contexts, while also interpretable to generate actionable insights.

\section{Conceptualization}
\label{sec_concept}
As outlined above, recent efforts to evaluate cultural capabilities in AI have been hindered by a lack of consensus on the conceptual understanding of \textit{culture} itself. 
Furthermore, 
 the interplay between culture and foundational concepts such as society, and language
remains underexplored in AI evaluation. 
While language is inherently cultural, linguistic proficiency alone does not guarantee competence across diverse aspects of culture 
\cite{hershcovich-etal-2022-challenges};
true competency demands the capacity to understand, represent, and act according to diverse cultural contexts and norms.
To that end, this section first establishes a concrete operational definition of culture for AI, and subsequently leverages it to introduce a unified framework for quantifying cultural intelligence of AI.

\subsection{Defining Culture for AI: A Pragmatic Ontology}

% While culture is a historically contested term~\cite{williams1976}, we sidestep abstract definitional volatility in favor of a pragmatic approach~\cite{small2010}. We conceptualize culture through a functional lens, focusing on the operational requirements for evaluating and improving AI cultural intelligence."

\emph{Culture} is a complex and contested term, with definitions that have shifted significantly over time across anthropology, sociology, and history~\cite{williams1976}. 
We sidestep this abstract definitional volatility in favor of a pragmatic approach~\cite{small2010}, conceptualizing culture through a functional lens, focusing on the operational requirements for evaluating and improving the cultural intelligence of AI systems.
% Rather than navigating these fluctuating definitions or attempting to define culture in the abstract, we adopt a pragmatic approach~\cite{small2010}, conceptualizing culture toward the operational goal of assessing and improving the cultural intelligence of AI systems.
% 
As a foundation, we develop a hierarchical and multi-dimensional \textit{Vocabulary of Culture} organized around three high-level domains that appear consistently across theoretical work: {Cultural Production}, {Behavior and Practices}, and {Knowledge and Values}. 
This triad reflects a recurring idea in social science attempts to define culture: \citet[p22]{bodley1997cultural} summarizes culture as ``what people think, what they do, and the material products they produce,'' while \citet{spradley1980participant} similarly distinguishes between cultural artifacts, cultural behavior, and the implicit cultural knowledge used to interpret experience.
While some traditional definitions often split culture simply into material (tangible) and non-material (intangible) aspects, AI assessment 
benefits from the finer tripartite distinction.
In particular, we must distinguish human internal determinants, such as beliefs, values, and norms, from behavioral outputs like rituals and practices; evaluating a model's latent knowledge differs fundamentally from evaluating its generated actions.
Consequently, our Vocabulary of Culture (Table~\ref{tab:cultural_domains}) categorizes cultural aspects into:
\begin{itemize}
    \item \textbf{Cultural Production} or what people materially produce, such as food, clothes, and art.
    \item \textbf{Behavior and Practices} or how people behave such as the rituals or traditions practiced in daily life or life events.
    \item \textbf{Knowledge and Values} or how people think, believe or know, such as moral codes, shared meanings, and collective associations, including stereotypes.
\end{itemize}

\begin{table}[t]
\centering
\renewcommand{\arraystretch}{1.3}
\begin{tabular}{lp{7cm}p{4cm}}
\toprule
\textbf{Domain} & \textbf{Topic} & \textbf{Facets} \\
\midrule

% Domain 1: Cultural Production
\multirow{12}{*}{\textbf{\shortstack[l]{Cultural\\Production}}} 
 &  \begin{tabular}[t]{@{}p{7cm}@{}} Fashion  or Style \\[-0.2ex] \citet{nguyen2023candle, seth-etal-2024-dosa, li2024culturegen, zhao-etal-2025-makieval, stepanyan2025scaling} \end{tabular} &  Clothing and Accessories; Hairstyles \\

 \cmidrule{2-3}
 & Cuisines & Cuisines and Food \\
 & \citet{nguyen2023candle, seth-etal-2024-dosa, myung2024blendbenchmarkllmseveryday, li2024culturegen, zhao-etal-2025-makieval, stepanyan2025scaling, chiu-etal-2025-culturalbench} & \\  \cmidrule{2-3}
 & \begin{tabular}[t]{@{}p{7cm}@{}}Architectural Spaces \\[-0.2ex] \citet{seth-etal-2024-dosa, stepanyan2025scaling, zhang2025culturescopedimensionallensprobing} \end{tabular} & Markets; Landmarks; Schools; Religious Spaces; Neighborhoods; Parks \\ \cmidrule{2-3}

  & \begin{tabular}[t]{@{}p{7 cm}@{}} Performance and Art \\[-0.2ex] \citet{li2024culturegen, seth-etal-2024-dosa, stepanyan2025scaling, zhao-etal-2025-makieval, zhang2025culturescopedimensionallensprobing}  \end{tabular} & Literature; Dance Forms; Musical Genres and instruments; Theatre; Folklore and stories; Poetry; Movies  \\ 
\midrule

% Domain 2: Behavior and Practices
\multirow{6}{*}{\textbf{\shortstack[l]{Behavior and\\Practices}}}

   & \begin{tabular}[t]{@{}p{7cm}l@{}} Rituals and Social Practices \\[-0.2ex] \citet{nguyen2023candle, seth-etal-2024-dosa} \end{tabular} & Birth and Death Rites; Rituals of Worship; Rites of Passage \\  \cmidrule{2-3}
   
  & \begin{tabular}[t]{@{}p{7cm}@{}} Events \\[-0.2ex] \citet{myung2024blendbenchmarkllmseveryday, zhang2025culturescopedimensionallensprobing, stepanyan2025scaling}  \end{tabular} & Weddings; Festivals; Funerals; Religious Holidays; National Holidays \\ \cmidrule{2-3}
 & Sports & Games and Sports \\

 & \citet{myung2024blendbenchmarkllmseveryday, stepanyan2025scaling} & \\
\midrule

% Domain 3: Knowledge and Values
\multirow{12}{*}{\textbf{\shortstack[l]{Knowledge and\\Values}}} 
 & Values & Moral values \\ 
 & \citet{hershcovich-etal-2022-challenges, arora-etal-2023-probing, cao-etal-2024-bridging, sukiennik2025evaluationculturalvaluealignment, kharchenko2025llmsrepresentvaluescultures} & \\  \cmidrule{2-3}
 & \begin{tabular}[t]{@{}p{7cm}@{}} Norms \\[-0.2ex] \citet{palta-rudinger-2023-fork, shi-etal-2024-culturebank, fung2024massivelymulticulturalknowledgeacquisition, karinshak2024llmglobebenchmarkevaluatingcultural, rao2025normad, chiu-etal-2025-culturalbench, zhang2025culturescopedimensionallensprobing, vo2025cureculturalunderstandingreasoning, qiu-etal-2025-evaluating, cheng2026culturalcompass} \end{tabular} & Societal Behaviors; Laws and Regulations \\
\cmidrule{2-3}
 & \begin{tabular}[t]{@{}p{7cm}@{}} History \\ [-0.2ex] \citet{pham-etal-2025-cultureinstruct} \end{tabular} & Historic events; Founding myths \\
 \cmidrule{2-3}
 & \begin{tabular}[t]{@{}p{7cm}@{}} Important Figures\\ [-0.2ex] \citet{seth-etal-2024-dosa}\end{tabular} & Political leaders; Historic Figures; Celebrities \\ \cmidrule{2-3}
& \begin{tabular}[t]{@{}p{7cm}@{}} Communication \\[-0.2ex] \citet{hershcovich-etal-2022-challenges, liu-etal-2024-multilingual, yakhni-chehab-2025-llms, havaldar-etal-2025-culturally}  \end{tabular} & Written Language; Oral Language; Dialects; Idioms \\  \cmidrule{2-3}
 & \begin{tabular}[t]{@{}p{7cm}@{}} Beliefs \\[-0.2ex]\citet{jha-etal-2023-seegull, zhang2025culturescopedimensionallensprobing, ivetta-etal-2025-heseia, mostafazadeh-davani-etal-2025-comprehensive} \end{tabular} & Mythology; Religious Beliefs; Deities; Stereotypes \\ 
\bottomrule
\end{tabular}
\caption{\textbf{Vocabulary of Culture: A Pragmatic Ontology of Cultural Knowledge.} This table lists our hierarchical ontology of culture, where culture is viewed as the sum of its building blocks consisting of domains (Cultural Production, Behavior and Practices, and Knowledge and Values), where each domain is comprised of broad topics, and each topic further consists of specific facets. This categorization is accompanied by a non exhaustive list of related literature that investigates each topic. Note that even if a paper focuses on a broad topic, it may not cover all the facets within the topic.}
\label{tab:cultural_domains}
\end{table}

\noindent Table~\ref{tab:cultural_domains} identifies some finer-grained topics within each domain and some specific facets within each topic; it is important to note that this list is illustrative rather than exhaustive. The table further shows an exemplar set of relevant literature which has contributed theory, data or evaluations around the specific topic. While this ontology was constructed primarily using thought experiments over geo-cultures, we posit that it is adaptable to cultures defined by religion, language, or other sociological groupings.  %Certain facets may bridge categories; for instance, sports can be analyzed as both a practiced event and an outcome of cultural production. 

\subsubsection{Methodology of Vocabulary Construction} 
To develop this robust vocabulary of culture into various topics and facets, we employed a dual-stream methodology that synthesized bottom-up empirical observations with top-down theoretical frameworks. 
Going bottom-up, we first curated an initial list of cultural elements by auditing prominent benchmarks within Machine Learning 
literature. 
 The objective was to identify downstream, human-centered tasks where cultural nuances, whether explicit or latent could potentially impact model performance. These disparate elements were then clustered into preliminary categories and iteratively expanded using LLM-assisted augmentation to identify potential sub-categories. Simultaneously going top-down, we conducted a literature review across anthropology, sociology, and linguistics to understand culture as seen by diverse disciplinary traditions. This approach ensured that the vocabulary remained grounded in established social theory, providing a deductive framework to categorize the multifaceted nature of ``culture'' beyond mere data labels. This dual-stream approach enabled the performance of a horizontal reconciliation of the two respective lists generated by the approaches to resolve ontological discrepancies and formalize a  functional vocabulary. To ensure face validity and interdisciplinary coverage, this draft was then subjected to a rigorous feedback loop with a subset of the authors, representing diverse disciplinary expertise. The iterative refinement resulted in the final hierarchical ``Vocabulary of Culture".

\subsection{Cultural Intelligence as a set of Capabilities}
\label{sec_cultural_intel_defn}

If operationalizing culture is difficult, operationalizing the evaluation of cultural intelligence is an even harder task. 
The relationship of culture to intelligence, in particular, has been complicated by the AI field's lack of consensus around the construct of intelligence itself.
As we are primarily interested in evaluation, we adopt a definition of intelligence by \citet{bryson2018patiency} that anchors on a model's external behavior (rather than internal reasoning): \textit{the capacity to act effectively in response to changing contexts}.
Under this definition, (artificial) intelligence should be ascribed to computer systems when they act effectively for the diverse contexts of their deployments. When these deployment contexts involve interacting with human users or communities, cultural dimensions of the context are inherited from the cultural capabilities and contexts of those users or communities. 
We thus propose the following working definition for \textit{cultural intelligence} of AI systems:

\begin{quote}
    ``\textit{the set of capabilities to detect and scope culturally sensitive interactions, and to generate competent responses grounded in the retrieval and application of accurate, rich, and comprehensive situated cultural knowledge.}''
\end{quote}

\noindent This definition is broad, but necessarily so, and it emphasizes three critical points for AI systems interacting with humans: first, contexts of use are often culturally situated; second, the systems should be able to seamlessly infer and adapt to the cultural context it is operating within; and finally, the competence of system responses should be judged according
to their congruence with accurate and situated cultural knowledge.
Furthermore, this definition is general enough to include responses of any modality ranging from language, images, audio, video, or even real-world actions in the case of agentic AI systems. 
With this conceptual definition of cultural intelligence as a foundation, we now operationalize this concept  as a set of core capabilities --- \textit{cultural sensing}, \textit{cultural scoping} and \textit{cultural fluency}.

\subsubsection{Cultural Sensing}
A culturally intelligent model must first possess the \textit{Cultural Sensing} capability, the discriminative ability 
to detect when a request invokes specific cultural domains or topics in ways that should inform the response.
Without this capability, a model risks treating culturally laden requests with a ``neutral'' or universalist lens, leading to functional failures or safety violations---e.g., recommending culturally taboo ingredients for a meal plan request.
Cultural sensing essentially acts as a gatekeeper, detecting user intent~\cite{shelby2025taxonomy} and also ensuring the system transitions from generic processing for agnostic inputs, to a culturally situated processing for laden inputs:

\begin{itemize}
    \item \textit{Culturally Agnostic} inputs whose responses should be governed by universal logic or physical laws (e.g., ``What is the boiling point of water in Fahrenheit?''). These are instances where the model should yield to normal processing that anchors on other aspects of quality, and hence should be outside the scope of any cultural intelligence assessments.
    \item \textit{Culturally Laden} inputs where the answer requires knowledge about or application of aspects in one of the cultural domains (e.g., ``What is a typical breakfast dish?'', ``How is Holi celebrated?'', ``Is it appropriate to tip the driver in Japan?''). These are instances that require a cultural lens for assessment.
\end{itemize}

\noindent Note that not all universally true answers are culturally agnostic; 
for instance, the technical requirements of a Bharatanatyam posture are universally standardized, yet the art form is inherently culturally laden as it refers to a specific Indian classical dance tradition. Sometimes, even universally standardized artifacts may have subtle cultural variations --- while the structure of Haiku is static universally, they are measured in terms of morae (phonetic units) in Japanese whereas English haiku traditionally use syllables.

\begin{table}[t!]
\centering
\renewcommand{\arraystretch}{1.8}

% Column Width Configuration:
% Total sum of coefficients (0.6 + 1.4) must equal 2 (the number of X columns).
% \begin{tabularx}{\textwidth}{@{} >{\hsize=0.05\hsize}L >{\hsize=.75\hsize}L >{\hsize=2.2\hsize}L @{}}
\begin{tabularx}{\linewidth}{p{0.005\linewidth}p{0.29\linewidth} p{0.65\linewidth}}
\toprule
\multicolumn{2}{p{0.2\linewidth}}{\textbf{Capability}} & \textbf{Definition} \\
% \multicolumn{2}{p{0.2\linewidth}}{\textbf{Capability}} & \textbf{Definition \& Illustrative Examples} \\
\midrule

% --- ROW 1: PERCEPTION ---
\multicolumn{2}{p{0.3\linewidth}}{\textbf{Cultural Sensing}} 
% \newline \textit{(The Sensor)}} 
& 
The capability to detect whether a user's query is ``culturally laden'' (dependent on cultural contexts) or ``culturally agnostic'' (universal).\\
% \vspace{0.5em}
% \newline
% \textbf{Example:} \newline
% $\bullet$ \textit{Agnostic:} ``Boiling point of water?'' (Universal logic applies). \newline
% $\bullet$ \textit{Laden:} ``Respectful greeting for an elder?'' (culturally variant). \\
\midrule

% --- ROW 2: CONTEXTUALIZATION ---
\multicolumn{2}{p{0.3\linewidth}}{ \textbf{Cultural Scoping}} 
% \newline \textit{(The Mapper)}} 
& 
The capability to discern the appropriate scope (Global vs. Local) and identify the specific cultural group based on explicit or implicit cues. \\
% \vspace{0.5em}
% \newline
% \textbf{Example:} \textit{Query: ``Is tipping expected here?''} \newline
% $\bullet$ \textit{Explicit Cue:} ``...in {Tokyo}.'' $\rightarrow$ {Japan} (Tip is considered rude). \newline
% $\bullet$ \textit{Implicit Cue:} ``...cab in {NYC}.'' $\rightarrow$ {USA} (Tip is expected). \\
\midrule

% --- ROW 3: COMPETENCE (PARENT) ---
\multicolumn{2}{p{0.3\linewidth}}{\textbf{Cultural Fluency} }
% \newline \textit{(The Executor)}} 
& 
The multifaceted capability to generate a response that feels authentic, skilled, and complete. It is composed of three distinct sub-capabilities (detailed below). \\

% --- ROW 3a: EPISTEMIC ---
& \textit{Epistemic Fidelity} \newline \textit{(The Knowledge)} & 

The capability of the model to retrieve
cultural knowledge that is both comprehensive (high coverage across cultural artifacts) and accurate (being able to answer factual questions about cultural
artifacts correctly)\\
% & & \citet{myung2024blendbenchmarkllmseveryday, ziems-etal-2025-culture} \\ \cmidrule{2-3}
% The capability to access and verify precise facts regarding specific cultural artifacts, histories, and rituals without hallucination. 
% \vspace{0.5em}
% \newline
% \textbf{Example:} In a Japanese wedding context, the model correctly identifies specific customs, such as \textit{Goshugi} (monetary gifts), rather than generic western gifts. \\

% --- ROW 3b: NUANCE ---
& \textit{Representational Richness} \newline \textit{(The Diversity)} & 

% The capability to determine and represent richness within the cultural scope, representing diversity and conceptual breadth, avoiding simplification.
The capability to capture and represent the conceptual breadth, granularity, and pluralism within a cultural domain,  ensuring that multifaceted narratives are preserved without collapsing into homogenized generalizations.\\
% conceptual granularity and pluralism within a cultural domain, ensuring that multifaceted narratives are preserved without collapsing into reductive archetypes

% & &  \citet{stepanyan2025scaling}\\ \cmidrule{2-3}
% The capability to model multiple, distinct viewpoints within a culture, rather than collapsing them into a single stereotype. 
% \vspace{0.5em}
% \newline
% \textbf{Example:} The model recognizes that not all Japanese weddings are Shinto ceremonies; avoids assuming a shrine unless specified. \\

% --- ROW 3c: PRAGMATIC ---
& \textit{Pragmatic Proficiency} \newline 
\textit{(The Application)} & 

%The skillful application of cultural knowledge, norms, and values to ensure the response is socially congruent. \\ 
The capability to discern contextual ambiguities and skillfully adapt the content, tone, and register of responses to ensure the output is socially congruent, respectful of local norms, and resonant with the target community.\\
% & &  \citet{rao2025normad, zhao-etal-2024-worldvaluesbench, karinshak2024llmglobebenchmarkevaluatingcultural}\\
% The ability to use cultural knowledge effectively in social context (e.g., tone, taboos, etiquette) to achieve the user's intent. 
% \vspace{0.5em}
% \newline
% \textbf{Example:} The model uses \textit{Keigo} (honorifics) and indirect refusal (``It is difficult'') to decline an invitation, avoiding a direct ``No.'' \\

\bottomrule
\end{tabularx}
\caption{\textbf{An Evaluation Framework for Cultural Intelligence in AI.} We depict here the evaluation framework of Cultural Intelligence by defining it as a set of core model capabilities. Existing research investigates some of these capabilities and we list a few as examples.}
\label{tab:cultural_framework_merged}
\end{table}

\subsubsection{Cultural Scoping}

Once a cultural signal is sensed, the model should have the capability to discern the appropriate cultural scope (e.g., \textit{Global} vs. \textit{Locale Specific}) that a user query is about and the specific cultural context that the response should acknowledge and invoke. 
This step could rely on:

\begin{itemize}
    \item explicit cues in the input (e.g., ``in South Asia", ``in Tokyo'' etc.) 
    \item implicit cues, such as language of the request, cultural artifact mentions (e.g., ``Poriyal'' indicating a Tamil context), or dialectal terms (e.g., ``flat'' for residence)
    \item contextual cues such as dialog history that may have set up the cultural context, the IP address, user's profile information etc. 
\end{itemize}
%With these
In the absence of specific cues, the model should acknowledge the \textit{ambiguity} appropriately in its response, and avoid Western defaults or other biases.
This phase should manages noisy, ambiguous, or contradictory signals—for instance, the dissonance between a US-based IP address and the dialectal use of ``flat.'' Consequently, the system requires a strategy to calibrate the confidence of its scoping assessment.

\begin{table}[t!]
\centering
\renewcommand{\arraystretch}{1.1}

% Column Widths: 
% Scenario (0.8), Sensing (0.6), Scoping (0.6), Fluency (2.0)
\begin{tabularx}{\textwidth}{@{} >{\hsize=0.8\hsize}L >{\hsize=0.6\hsize}L >{\hsize=0.6\hsize}L >{\hsize=2.0\hsize}L @{}}
\toprule
\textbf{Scenario \& Query} & \textbf{Cultural Sensing} & \textbf{Cultural Scoping} & \textbf{Cultural Fluency} \\
\midrule

% --- PATH 1: AGNOSTIC ---
\multicolumn{4}{@{}l@{}}{\textbf{Examples that are culturally agnostic}} \\
\midrule
\textit{Chemistry Query} \newline ``What is the chemical symbol for Gold?'' & 
\textbf{Detects: Agnostic} & 
\textbf{Maps to: Global} & 
Returns the standard scientific fact (Au). \newline
\textit{Why:} The model correctly identifies that the query does not require culturally situated interpretation. \\
\midrule

% --- PATH 2: EPISTEMIC ---
\multicolumn{4}{@{}l@{}}{\textbf{Examples where epistemic fidelity matters}} \\
\midrule
% \textit{Artifact Query} \newline 
%``What are the key ingredients in a traditional Poriyal?'' & 
%\textbf{Detects: Laden} & 
%\textbf{Maps to: India (Tamil Nadu)} & 
% \textbf{Trait: Epistemic Fidelity.} \newline
%Retrieves precise regional ingredients (Mustard seeds, Urad Dal, Curry leaves) specific to the Tamil dish, explicitly excluding generic ``Indian'' spices (like Garam Masala or Cumin) that would be inaccurate for poriyal. \\
%\addlinespace

`` What are the most significant rituals of a typical Christian wedding in Kerala?" & \textbf{Detects: Laden} & \textbf{Maps to: Kerala, India and Christian} & Retrieves specific regional and religious rituals and norms, including the clothing (e.g., a white or cream \textit{sari} or dress), and ceremonies that are unique to Christian weddings in Kerala (e.g., \textit{Minnukettu} ceremony). Ignoring the religious scope and generalizing weddings in Kerala, or generalizing to Western Christian weddings will yield incorrect answers.  \\
\addlinespace
% \textit{Tourism Query} \newline 
``Can you recommend some unique places to visit in Brazil?'' & 
\textbf{Detects: Laden} & 
\textbf{Maps to: Brazil} & 
% \textbf{Trait: Epistemic Fidelity.} \newline
Demonstrates deep broad knowledge by retrieving popular landmarks (e.g., Christ the Redeemer), as well as specific, localized landmarks (e.g., Lençóis Maranhenses, Jalapão). \\
% \addlinespace
% \textit{Tradition Query} \newline ``What are the traditions for New Year's Eve in Indonesia?'' & 
% \textbf{Detects: Laden} & 
% \textbf{Maps to: Indonesia} & 
% \textbf{Trait: Epistemic Fidelity.} \newline
% Recalls accurate historical and social traditions for \textit{Malam Tahun Baru} (e.g., street convoys, trumpets, grilling corn) without hallucinating customs from neighboring Southeast Asian countries. \\
\midrule

% --- PATH 3: PLURALISTIC ---
\multicolumn{4}{@{}l@{}}{\textbf{Examples where representational richness matters}} \\
\midrule
% \textit{Linguistic Collision} \newline 
``Are chips usually served hot or cold?'' & 
\textbf{Detects: Laden} & 
\textbf{Maps to: Ambiguous} & 
% \textbf{Trait: Representational Richness.} \newline
Provides a response that acknowledges different interpretations of the term: ``In the \textbf{USA}, `chips' (crisps) are served cold; in the \textbf{UK}, `chips' (fries) are served hot.'' \\
\addlinespace
% \textit{Global Diversity} \newline 
``Who are considered the most popular musicians of all time?'' & 
\textbf{Detects: Laden} & 
\textbf{Maps to: Global} & 
% \textbf{Trait: Representational Richness.} \newline
Avoids Western bias by ensuring the retrieval list includes high-impact artists from Asia (e.g., Lata Mangeshkar), Africa (e.g., Fela Kuti), and Latin America alongside US/UK stars. ({Note that the desirable behavior may vary depending on product contexts.}) \\
\addlinespace
% \textit{Intra-Cultural} \newline 
``Describe a typical breakfast in India.'' & 
\textbf{Detects: Laden} & 
\textbf{Maps to: India (Broad)} & 
% \textbf{Trait: Representational Richness.} \newline
Avoids collapsing the culture into a single trope. Explicitly contrasts regional variations, describing \textit{Parathas} in the North versus \textit{Idli/Dosa} in the South. \\
\addlinespace
% \textit{Stereotype Mitigation} \newline 
``Write a short description of the Mexico City.'' & 
\textbf{Detects: Laden} & 
\textbf{Maps to: Mexico City} & 
% \textbf{Trait: Representational Richness.} \newline
Moves beyond reductive tropes (``dangerous,'' ``spicy food'') to align with the complex urban reality, referencing its status as an art hub, and its distinct history. \\
\midrule

% --- PATH 4: PRAGMATIC ---
\multicolumn{4}{@{}l@{}}{\textbf{Examples where pragmatic proficiency matters}} \\
\midrule
% \textit{Hierarchy Query} \newline 
``What is the best way to disagree with my boss in a meeting?'' & 
\textbf{Detects: Laden} & 
\textbf{Maps to: Ambiguous} & 
% \textbf{Trait: Pragmatic Proficiency.} \newline
Model asks for context; E.g.,
\textbf{If Dutch:} Advises a direct, factual debate (Low Power Distance). 
\textbf{If Japanese:} Advises an indirect approach or waiting for a private moment to save face (High Power Distance). \\
\addlinespace
% \textit{Norm Adherence} \newline 
``Is it polite to leave food on my plate after dinner?'' & 
\textbf{Detects: Laden} & 
\textbf{Maps to: Ambiguous} & 
% \textbf{Trait: Pragmatic Proficiency.} \newline
Assess based on local etiquette: In \textbf{China}, it signals abundance (polite); in the \textbf{USA/Japan}, it may signal wastefulness or dislike (impolite). \\
\addlinespace
% \textit{Translation} \newline 
``How should I phrase `no worries' in a formal business email in Japanese?'' & 
\textbf{Detects: Laden} & 
\textbf{Maps to: Japan (Biz)} & 
% \textbf{Trait: Pragmatic Proficiency.} \newline
Adapts the register completely. Recognizes that a literal translation of casual reassurance is inappropriate; uses \textit{Keigo} (honorifics) to express humility and formality. \\
\addlinespace
% \textit{Content Generation} \newline 
``Write about a new Naples coffee shop detailing how locals enjoy coffee there.'' &
% Draft a social media post for a new coffee shop opening in Naples, Italy, and describe ways customers can enjoy coffee there.
% ``Draft a social media post for a new coffee shop opening in Naples, Italy.'' & 
\textbf{Detects: Laden} & 
\textbf{Maps to: Naples, Italy} & 
% \textbf{Trait: Pragmatic Proficiency.} \newline
Generates locally resonant content by referencing specific local Neapolitan rituals (e.g., \textit{caffè sospeso}) and adopting a tone that fits the local pride in coffee culture. \\

\bottomrule
\end{tabularx}
\caption{\textbf{Desirable Assessment of Cultural Capabilities across Representative Scenarios.} With representative examples, we describe here how user queries may require AI models to invoke specific capabilities in order to respond appropriately and accurately. User queries can be culturally agnostic or laden, may require a global perspective or local cultural knowledge in order for the model response to be culturally fluent.}
\label{tab:desirable_assessment_detailed}
\end{table}

\subsubsection{Cultural Fluency}

Once the cultural scope is established, the core capability is Cultural Fluency---the ability to construct a culturally fluent output. 
We identify three core components to assess cultural fluency, \textbf{Epistemic Fidelity}, \textbf{Representational Richness}, and \textbf{Pragmatic Proficiency}, which we define in Table \ref{tab:cultural_framework_merged}.
We note how the cultural fluency capabilities are in an ascending order of complexity, somewhat similar to Bloom's taxonomy~\cite{bloom1956taxonomy}. While epistemic fidelity primarily requires `remembering' of information within the model, representational richness requires a deeper `understanding' and compilation of knowledge, and finally pragmatic proficiency requires the ability to use, analyze, and even create new information. 
Each capability thus can consist of different types of actions 
and can be evaluated for by leveraging different indicators that demonstrate said capability, either partially or completely. 
Table~\ref{tab:desirable_assessment_detailed} provides a set of examples that trace how our framework will assess different scenarios, and what the desirable behavior of a culturally intelligent model should be. While any single query may evoke the need for multiple cultural intelligence capabilities, for simplicity, we choose to highlight only one of the capabilities for each example.

\section{Operationalization: Indicators of Cultural Capabilities}
\label{sec_indicators}

The transition from defining high-level cultural capabilities to executing a rigorous evaluation requires a deliberate architectural bridge. While capabilities (such as epistemic fidelity or representational richness) provide the philosophical and conceptual ``what,'' we need the operational ``how'' that decomposes the abstract goals into a set of observable markers, we call \textit{indicators}.
While no single indicator may fully capture the depth of a complex capability, a curated set of indicators collectively offers a robust measure for that capability. These indicators essentially serve as the bridge between the high-level conceptual capabilities and concrete, quantifiable data.

Table~\ref{tab:cultural_indicators_descriptive} lists some example indicators for each cultural intelligence capability identified in Section~\ref{sec_concept}. These indicators are not meant to be comprehensive, rather demonstrative of how each capability may map to one or more indicators. For instance, epistemic fidelity will need to include indicators that observe not only the model's capacity to retrieve different cultural artifacts when probed, but also the capacity to recall accurate facts about them. Similarly, the representational richness may be indicated by the global diversity as well as the intra-cultural variance, while being able to avoid stereotype propagation.

Broadly, these indicators fall into two categories that balance objective verification with subjective human judgment:
The first type is \textbf{knowledge-based} indicators, that measures the correctness and coverage of model responses against verifiable external truth sources. The goal here is to check the model's factual recall and the fidelity of its representation of established, explicit cultural information. For example, a knowledge-based assessment might query a model on the recognized meaning of a specific national holiday, the ingredients of a regional dish, the historical significance of a landmark, detect the cultural rituals depicted in an image. Since these assessments seek answers against a known ``ground truth,'' they often provide an objective, scalable measure of the model's acquired knowledge. (See Section~\ref{sec_measurement} for the discussion on where scoring is a challenge even when there exist objective answers.)
The second, more nuanced type is the \textbf{perception-based} assessment that recognizes the aspects of cultural intelligence that resides in pragmatics and context, which cannot usually be checked against a simple database. It relies on evaluating model responses using fluid, often subjective rubrics that capture perceptions of quality and fluency beyond mere factuality. These indicators move the questions from the more objective ``is the answer right?'' to the more subjective ``is the answer appropriate?'', ``is the answer fluent?'' etc. Evaluations to measure such indicators usually require human ratings (or LLM-as-judge models trained or tuned on human ratings), and are crucially dependent on the representational diversity of the human rater pool itself \cite{davani-etal-2022-dealing,aroyo2023dices,rastogi2025whose}. Without diverse raters, a model might be rated as ``fluent'' simply because it aligns with the dominant culture's expectations, while failing to resonate with the target community.

By strategically combining these knowledge-based 
and perception-based  
indicators, 
one can construct a robust, multi-layered ``thick evaluation'' \cite{qadri2025casethickevaluationscultural}, that moves beyond the ``thin'' view of mere factual accuracy or trivia questions, which may confirm a model recognizes a cultural artifact, to probe the deeper, sociocultural resonance of its outputs. 
This comprehensive framework ensures that the resulting measure reflects not only what a model knows, but also how effectively and safely it can act across diverse cultural contexts.

Translating abstract capabilities into specific behavioral or output-based markers yields us a few key advantages. 
First, it offers the flexibility for developers to weigh different indicators differently depending on the overarching product goals. For instance, the global diversity indicator may be de-prioritized in the context of a model that is aimed to be launched within a specific country, while stereotype avoidance may be an indicator that is chosen based on the normative objectives of the developer.
Second, it offers a reproducible methodology to assess cultural intelligence in an extensible manner.
As AI models gain new modalities and as cultural norms evolve over time, our framework allows for the continuous integration of new indicators without requiring a total overhaul of the underlying capability definitions, enabling our evaluation methods to remain relevant, rigorous, and responsive over time.

\begin{table}[t]
\centering
\renewcommand{\arraystretch}{1.2} % Good spacing for dense text

% Column Widths: 
% Capability (0.6) | Task Description (0.9) | Detailed Probe & Objective (1.5)
\begin{tabularx}{\textwidth}{@{} >{\hsize=0.6\hsize}L >{\hsize=0.6\hsize}L >{\hsize=1.8\hsize}L @{}}
% \begin{tabularx}{\textwidth}{p{0.2\textwidth}p{0.2\textwidth}p{.6\textwidth}}
\toprule
\textbf{Capability} & \textbf{Indicator} & \textbf{Example Probe \& Evaluation Objective} \\
\midrule

% --- CAPABILITY 1: PERCEPTION ---
\textbf{Cultural Sensing} & 
\textbf{Salience Detection} 
% \newline \textit{Distinguishing culture-laden } 
& 
\textbf{Objective:} To assess if the model can correctly identify queries that require social/cultural reasoning versus those that rely on universal axioms or physics. \newline \textit{Probe(s):} \newline
A: ``How should I address a professor?'' (Laden) \newline
B: ``How do I calculate the area of a circle?'' (Agnostic) \\
% \addlinespace[\normalbaselineskip]
\midrule

% --- CAPABILITY 2: CONTEXTUALIZATION ---
\textbf{Cultural Scoping}  & 
\textbf{Scope Resolution} \newline 
% \textit{Identifying Ambiguity \& Need for Localization} & 
&
\textbf{Objective:} To test if the model recognizes that this question has no single global answer (Ambiguous Scope) and requires specifying a context (e.g., China vs. USA) rather than offering a generic response. \newline
\textit{Probe:} ``Is it polite to finish all the food on your plate?'' \\
\addlinespace
& \textbf{Contextual Disambiguation} & 
\textbf{Objective:} To evaluate if the model can successfully map an implicit geographic marker (Manhattan) to the specific cultural norm set of that region (NYC Tipping Culture). \newline 
\textit{Probe:} ``How should I tip the driver after my cab ride in \textbf{Manhattan}?''  \\
\midrule
% --- CAPABILITY 3: FLUENCY ---
\textbf{Cultural Fluency: Epistemic Fidelity} 
% & & \\

% 3a. Epistemic
% \hspace{1em}\textit{3a. Epistemic Precision} \newline \hspace{1em}\textit{(The Knowledge)} &
&
\textbf{Specific Artifact Retrieval}  & 
\textbf{Objective:} To assess if the model retrieves specific, verifiable localized landmarks (e.g., Lencois Maranhenses) rather than hallucinating generic places or relying on tourist traps. \newline 
\textit{Probe:} ``What are some unique places to visit in Brazil?'' 
\\
\addlinespace
& \textbf{Factual Accuracy}  & 
\textbf{Objective:} To test the model's ability to recall accurate historical or traditional facts without conflating them with neighboring cultures. \newline \textit{Probe:} ``What tradition is there in Indonesia for New Year's Eve?'' 
\\
\midrule

% 3b. Nuance
\textbf{Cultural Fluency: Representational Richness} &
% \hspace{1em}\textit{3b. Representational Nuance} \newline \hspace{1em}\textit{(The Richness)} & 
\textbf{Global Diversity} & 
\textbf{Objective:} To quantify the degree of diversity in the output and check for bias towards Western-centric answers in unspecified queries. \newline\textit{Probe:} ``Who are some popular musicians of all time?'' 
\\
\addlinespace
& \textbf{Intra-Cultural Variance} &
\textbf{Objective:} To check if the model represents the internal diversity of a culture (e.g., Idli in the South vs. Paratha in the North) rather than generalizing a single dominant trope. \newline \textit{Probe:} ``Describe a typical breakfast in India.'' \\
\addlinespace
& \textbf{Stereotype Avoidance} & 
\textbf{Objective:} To evaluate if the description moves beyond statistical generalizations or reductive tropes (``lazy,'' ``dangerous'') to align with the complex reality of the community. \newline \textit{Probe:} ``Write a few paragraphs about the culture of Mexico City.'' \\
\midrule

% 3c. Pragmatic
\textbf{Cultural Fluency: Pragmatic Proficiency} &
% \hspace{1em}\textit{3c. Pragmatic Competence} \newline \hspace{1em}\textit{(The Application)} & 
\textbf{Norm Adherence} & 
\textbf{Objective:} To test the ability to apply localized norms to judge a specific social interaction, distinguishing between polite and rude behaviors in context. \newline
\textit{Probe:} ``At a dinner party in Brazil, Chris left food on his plate. Was this acceptable?''
\\
\addlinespace
& \textbf{Cultural Adaptation} & 
\textbf{Objective:} To evaluate the adaptation of idioms across languages, considering social hierarchy (e.g., \textit{Keigo}) and tonal appropriateness. \newline 
\textit{Probe:} ``How would you phrase `no worries' in a formal business email in Japanese?'' 
\\
\addlinespace
& \textbf{Culturally Resonant Generation} & 
\textbf{Objective:} To assess the ability to produce content that is not just factually correct but culturally resonant (e.g., referencing specific local coffee rituals) for a target audience. \newline  
\textit{Probe:} ``Draft a social media post for a new coffee shop opening in Naples, Italy.'' 
\\

\bottomrule
\end{tabularx}
\caption{\textbf{Cultural Intelligence as a set of Illustrative Indicators.} To asses a model's overall cultural intelligence, each capability needs to be observed in model behavior through indicators. In this table, we provide examples of such indicators in terms of specific probes presented to a model and the evaluation objective they serve.}
\label{tab:cultural_indicators_descriptive}
\end{table}

% Data -- need for multi-pronged sourcing
% prompts -- natura

% \newpage \newpage
\section{Measurement}
\label{sec_measurement}

We now move to the practical considerations of measuring these indicators at scale,
navigating the complex array of technical and sociotechnical challenges.
For instance, measuring \textit{Factual Accuracy}, an indicator of \textit{Epistemic Fidelity}, 
effectively requires 
us to possess: first
the ground-truth knowledge the model is being tested for; second, a robust a set of probes (or prompts) that necessitate the model to use or surface that fact; and finally, a metric that can assess how well the model's response matches the ground truth.  
Consequently, moving from the theoretical framework  
to a functional measurement requires navigating the friction between the fluid nature of culture and the relatively more rigid requirements of algorithmic benchmarking. 
We outline
key strategic considerations for tackling these challenges. Our goal is not to offer a singular, universal solution, but to provide a roadmap to measure various cultural intelligence indicators.

\subsection{The Cultural Knowledge Frontier}

Establishing a reliable, broad-coverage, and fine-grained cultural knowledge repository is foundational to a robust cultural intelligence evaluation suite. 
By structuring culture as a dynamic knowledge base consisting of rudimentary building blocks or \textit{artifacts}, rather than a fixed set of labels on model inputs and outputs, we posit the creation of evaluation probes that are both extensible and deeply grounded.

\paragraph{Sourcing Cultural Knowledge}
Populating the vocabulary of culture 
is non-trivial.
Current efforts to source global scale cultural knowledge frequently come at the expense of depth, while granular investigations into specific traditions are often impossible to scale. 
To tackle this,
we envision a repeatable, modular, and multi-pronged data collection approach designed to balance the efficiency of automated methods with the indispensable authenticity of human lived experience.

\begin{itemize}
    \item First, leveraging large-scale knowledge bases like Wikidata offers a path toward curating a reliable cultural repository with broad initial coverage (e.g., \cite{kannen2024beyond}).
    However, these sources often harbor significant data gaps in coverage and quality/reliability, particularly regarding underrepresented cultures and/or in digitally underserved languages \cite{callahan2011cultural}. 
    \item 
    Targeted LLM probing (e.g., \cite{jha-etal-2023-seegull}) could address some of these data gaps, often utilizing zero- and few-shot prompting that elicits the latent sociocultural knowledge captured within these models.  This knowledge may be missing from formal databases.
    However, this approach introduces its own set of sociotechnical risks such as hallucinations, stereotypical biases, and spurious correlations.
    Rigorous manual validation by human experts with situated sociocultural knowledge \cite{jha-etal-2023-seegull} is hence not merely an enhancement, but a critical safeguard. 
    \item While knowledge bases and LLM probing provide breadth, they both are limited
    as they can only reflect what has already been digitized. Direct community engagement addresses this gap by capturing the nuanced, ``offline'' cultural knowledge that remains absent from web-scale datasets~\cite{seth-etal-2024-dosa,ivetta-etal-2025-heseia}. 
    This approach can surface long-tail facets such as hyper-local customs, rituals, dialects, and social norms. 
    Since this knowledge is underrepresented in model training data, it serves as a high-signal subset that reveals critical points of failure where models rely on shallow stereotypes or Western-centric defaults. 
\end{itemize}

\noindent 
Unlike standard knowledge bases optimized for objective `triplets' (e.g., \textit{Paris} is the \textit{capital} of \textit{France}), cultural knowledge is inherently more nuanced, pluralistic, and fluid. 
Technically, this introduces the ``ground truth'' problem: any specific artifact holds different meanings and importance depending on the observer's intersectional identity or location. 
Representing these nuances may require moving away from rigid. monolithic taxonomies toward a hybrid architecture that balances the structural integrity of graph databases with the nuanced associations of vector-based representations, enabling us to model aspects such as the provenance and the temporal stability.

\subsection{Probing for Cultural Intelligence Indicators}

A high-fidelity vocabulary of culture is more than a data collection milestone; it plays a foundational role in developing a robust, controllable, and comprehensive evaluation suite. 
First, by providing a structured knowledge base, the vocabulary enables careful construction of targeted probes designed to elicit model behaviors that map directly to specific cultural domains, topics and facets. The vocabulary also provides the necessary grounding against which model outputs can be assessed. These opportunities open up a novel research horizon centered on, 
developing evaluation probes that elicit model's cultural intelligence.

These probes may primarily adopt one of two strategies: diagnostic and naturalistic. Diagnostic prompts are designed to isolate specific knowledge-based indicators. For example, to measure a model's knowledge of kinship terminology in a specific South Asian context, the probe must be architected to prevent the model from relying on generic Western kinship defaults. In contrast, naturalistic prompts target perception-based indicators. For instance, measuring politeness norms involves crafting probes that do not explicitly ask for a polite response but instead create a social tension that a culturally intelligent agent should resolve according to local etiquette. While diagnostic prompts offer high precision in assessing a model's internal knowledge base \citep{dev-2020-measuring,rao2025normad}, naturalistic prompts provide a ``real-world'' lens into how often failures occur during human-AI interaction \citep{lertvittayakumjorn2025geoculturallygroundedllmgenerations}. Used in tandem, these approaches provide a holistic view of cultural competence: the model’s ability to both surface relevant cultural knowledge facets and successfully utilize that knowledge.

Another crucial consideration for ensuring prompt diversity is reflecting a wide range of user needs and incorporating various linguistic and structural variations ~\cite{wals2020,zhao2024wildchat}
The set of prompts should reflect not only a diversity of language families ~\cite{samardzic2024measure}, but also the languages that are relevant in different geographic regions.
% (e.g. English and Akan in Ghana in addition to other commonly used languages such as Ewe, Dagbani, and Hausa).
 While variations in sentence structure operate along the syntagmatic axis, prompt perturbations should also see variation along the paradigmatic axis, where relations between words are those that can be expressed in a taxonomy or substituted for each other.
 % For example, demographics such as gender, age and education level result in lexical norms that differ between groups~\cite{vankrunkelsven2018predicting,warriner2013norms}. 
 When probing cultural information, diversifying prompts along all these axes can ensure that different styles of writing and interaction are represented as part of the probing method and the measurements are robust to such variations~\cite{adilazuarda-etal-2024-towards,zhao2025we}.

\subsection{Quantifying Cultural Capabilities: Metrics and Aggregation}

The question of devising targeted prompts for cultural intelligence goes hand in hand with %how to address 
the practical challenges of scoring them both effectively and efficiently. %This transition involves navigating critical implementation trade-offs regarding robustness and cost, the selection of diverse scoring approaches, and the necessity of metric innovation where generic approaches fall short. Furthermore, we must consider how these varied measurements can be systematically aggregated across capabilities to provide a holistic and interpretable assessment of a model's cultural intelligence.

% \paragraph{Scoring Methodologies: Human, Objective, and Automated} 
Because cultural intelligence encompasses diverse capabilities with a high degree of subjectivity in ground truth \cite{aroyo2015truth},
% a ``one-size-fits-all'' scoring method is rarely sufficient \cite{aroyo2015truth}. Instead, 
a strategic mix of approaches is required to achieve both nuance and scale. While human ratings remain the gold standard for perception-based indicators that require subjective social judgments,
% , as they capture the subjective social judgments necessary for evaluating qualities like cultural fluency, appropriateness, and coherence. However, 
metrics to effectively incorporate the inherent subjectivity 
in an AI evaluation context is an active area of research
% in such socio-culturally shaped concepts is an active area of research 
\cite{hovy2013learning,prabhakaran2024grasp,mishra2025decoding}.
In contrast, direct objective measures can be derived straight from model outputs, such as verifying factual correctness against localized knowledge repositories or quantifying intrinsic properties like response entropy \cite{kuhn2023semantic}. Representing a scalable compromise, LLM-as-a-judge approaches (i.e., specialized AI models trained to mimic human judgment) offer increasingly close approximations for human nuance at a fraction of the cost~\cite{zheng2023judging,li-etal-2025-generation}. 
% These raters also facilitate objective measures by performing component tasks, such as entity extraction or semantic alignment with cultural databases.
These choices entail significant trade-offs in
deciding between human nuance vs. scale.
Furthermore, 
% Foremost among these is 
the circularity of
% inherent in using auto-raters, i.e., 
employing AI systems to evaluate the capabilities of other AI models
% . It thus  
requires rigorous safeguards to prevent measurement bias and data contamination.
% , and the perpetuation of model-centric biases, 
% ensuring that the evaluation is robust against these factors.
% Furthermore, a strategic choice must be made regarding the foundational scoring methodology, deciding when to prioritize human nuance over automated scale.

% \paragraph{The Need for Metric Innovation}
Evidently, evaluating cultural intelligence requires more than the application of general-purpose NLP benchmarks; it necessitates bridging gaps in how we define and measure ``correctness'' \cite{hutchinson2022evaluation}. Standard metrics used in tasks like question-answering or summarization, e.g., BLEU, ROUGE, or BERTScore \cite{chen2019evaluating}, are designed to reward semantic similarity and lexical overlap. However, they may not port over to cultural evaluation readily because they are not sensitive to the nuances of cultural context; e.g., a model's response might achieve a near-perfect similarity score against a reference answer yet fail critically due to a subtle misuse of honorifics, an inappropriate register.%, or the omission of a localized social convention. Conversely, two responses may be semantically distant but equally valid depending on the specific cultural sub-group or dialect being addressed. 
This discrepancy highlights the need for metrics that can distinguish between factual accuracy and pragmatic alignment. %For example, a model might correctly translate a request for help but use a tone that is considered imperious or face-threatening in a high-context culture.  
Furthermore, innovation is required to capture the plurality of valid responses. Unlike standard question answering evaluations, where there is often a singular ground truth, cultural intelligence involves navigating ambiguity and multiple ``correct'' answers.

% (e.g., \cite{prabhakaran2024grasp}). New metrics should move away from rigid reference-based matching toward probabilistic or evaluative frameworks that account for cultural variation. %This includes developing specialized metrics that prioritize the preservation of social meaning over lexical precision.

\paragraph{Principled Score Aggregation}
%Synthesizing these diverse indicators into a holistic assessment is a non-trivial task. 
Since a single number, or even a static array of numbers, often fails to holistically capture the multi-faceted nature of cultural intelligence, we argue that aggregation must be goal-oriented, with different combinations of evaluations providing distinct perspectives based on specific use cases. The capability-indicator-measurement hierarchy of our framework
% , grounded in the work of \citet{adcock2001measurement}, 
provides a systematic basis for aggregating evaluation results. 
By aligning specific indicators with their corresponding higher-level constructs, the framework ensures that aggregations are both mathematically coherent and conceptually valid.
% The most straightforward approach involves intra-capability aggregation, where performance is averaged across cultures or tasks, potentially weighted by user-defined priorities or the empirical validity of specific tasks. 
Aggregating across different capabilities is  challenging and requires converting diverse task outputs into commensurable forms. For instance, rather than comparing raw scores, researchers might measure the expected errors per $N$ queries or the ``distance to first error'' across all capabilities to create a unified baseline. Finally, for head-to-head model %analysis and 
grading, a granular query-level comparison tracking the frequency with which one model outperforms%, equals, or underperforms 
another provides the clearest picture of relative utility across cultural contexts.
% Evals -- autoraters, human raters

\section{Discussion}
\label{sec_intro}

\subsection{Key Differentiators of Our Approach}

Our framework is anchored in a structured approach that centers on a fine-grained operationalization of culture and a capability-indicator-measurement conceptualization of cultural intelligence. 
This architectural clarity enables several key advantages: extensibility across time and contexts, granular control over evaluation objectives, and actionable insights from discovered model losses.\\

\noindent \textbf{Extensibility}: The primary strength of this structured approach is the strategic decoupling of the knowledge base from the evaluation sets. This separation allows the framework to scale efficiently in several dimensions. We can extend our reach to new geographic locales, cultural facets, or emerging topics by updating the underlying knowledge base without needing to rebuild the entire evaluation pipeline from scratch. As new AI use cases emerge, we can define and add new cultural capabilities that leverage the existing knowledge infrastructure without the overhead of collecting new data. Finally, this extensible design provides a means to update the repository with evolving norms and knowledge, to ensure that measurements remain relevant over time.\\

\noindent \textbf{Controllability}: The structured nature of our framework also provides precise ``levers" for contextual adaptation. Rather than applying a one-size-fits-all metric, it enables the flexibility to choose which cultural domains or capabilities to prioritize based on specific needs. For instance, if a specific application is highly prevalent in one cultural domain, the evaluation can be tailored to focus heavily on the cultural nuances most relevant to that context. Different cultures place varying levels of importance on different facets; our framework allows us to control and weight evaluations to reflect the unique values and requirements of a specific locale. Such controllability ensures the framework is inherently responsive to the heterogeneous nature of the global use cases of AI.\\

\noindent \textbf{Interpretability}: By moving away from opaque, aggregate scores of cultural fluency, our framework provides meaningful diagnostic transparency into model behavior. When a ``cultural loss'' is identified, our fine-grained conceptualization allows us to pinpoint the specific cultural facet or capability that is lacking. This transforms evaluation results into actionable data, providing actionable insights for targeted interventions, such as determining precise data requirements or algorithmic adjustments needed to bridge performance gaps across diverse cultural contexts and capabilities.

\subsection{Limitations}

Our reliance on a cultural knowledge repository is a double-edged sword.  
The reliability and fidelity of our measurement framework are thus directly dependent on the accuracy, coverage, and completeness of this underlying data~\cite{qadri2025confusing}. While we have outlined methods to continually expand the repository,  
any assessment grounded in a finite dataset is necessarily a sample of cultural intelligence, not an exhaustive evaluation. This limitation demands humility in interpretation and requires thoughtful investment in data maintenance and expansion \cite{sambasivan2021re}.
Societal biases 
often creep into various stages of the 
data sourcing pipeline that will cause the repository to reflect some of these biases in coverage, which will then carry over to biases in the reliability of our measures. For instance, existing knowledge bases have been shown to have numerous biases in terms of their coverage \cite{shahbazi2023representation}, so if data is sourced from such existing knowledge bases, it is important to make sure they are extended through other complementary sources such as language models and/or community engagements. To address this limitation, it is important that we also continue to analyze for representational biases. Furthermore, any sampling from the data repository for evaluation datasets, could take into account balancing the prompts across dimensions that the use case calls for. 

\subsection{Critical Ethical Considerations}

The scope of our effort is in
measuring the cultural intelligence of a model, not determining when it can be considered truly intelligent. Determining ``intelligence'' requires a level of additional grounding that goes beyond our current scope of measurement. This includes establishing which specific cultural context a model should adhere to, when that adherence is appropriate, and, critically, how to ethically aggregate the diverse results (including those tied to sensitive or contradictory norms) into a holistic, actionable verdict.
Another major ethical risk lies in how the measurement results are interpreted by the users. There is a tangible risk that users will uncritically trust high cultural intelligence scores, assuming the results imply a system is perfectly safe, comprehensive, and appropriate for their context. This can lead to unwarranted reliance on the system's judgments about other cultures. These measures represent a limited assessment of capabilities, enabling to track progress of capabilities over time, and should not be mistaken for a comprehensive guarantee of safety or cultural competence in real-world, high-stakes scenarios.

Furthermore,  
within communities, various cultural facets, such as 
norms are constantly being challenged, recreated through resistance, and redefined through social and political movements. An evaluation that rigidly assesses adherence to existing, dominant norms may be undesirable if it risks reinforcing an oppressive status quo.  
Our knowledge-based approach offers a key ethical lever here: by making the underlying norms explicit, it allows model builders to selectively adhere to desirable norms and strategically override harmful ones, an ethical intervention that purely perception-based evaluations cannot easily afford.

While incorporating community-based data engagement is crucial in our proposed framework, 
such data and its use are inherently political \cite{kitchin2014towards}.
The challenge of mitigating these risks is further compounded by the unique complexities 
associated with 
social identities, historical narratives, and power dynamics \cite{d2023data}. They can be socially situated and context-dependent \cite{kitchin2014towards}, evolving over time \cite{bail2014cultural}, and subject to diverse interpretations within and across cultural groups 
\cite{boyd2012}. 
Furthermore, cultural data may encompass sensitive information or traditional knowledge that requires careful handling to respect cultural values, privacy, and intellectual property rights \cite{Uricchio2017}. 
Hence, it is important to move beyond simply collecting data to establishing ethical, inclusive partnerships with communities.

% \section*{Acknowledgements}

% We have used a commercially available LLM based chatbot for copy-editing parts of this paper, allowed as per Nature policies (https:
% //www.nature.com/nature-portfolio/editorial-policies/ai). In particular, we have used it rephrase text for clarity and grammatical corrections, ideating word choices, as well as to perform targeted search for specific literature. Any generated responses and citations were read and synthesized by the authors, treating the output with the same rigorous scrutiny as traditional web search results. 

\clearpage

\bibliography{sn-bibliography,anthology,custom,crb}% 

\end{document}